\documentclass[11pt]{article}

\usepackage[a4paper,margin=1in]{geometry}
\usepackage[T1]{fontenc}
\usepackage[utf8]{inputenc}
\usepackage{amsmath,amssymb}
\usepackage{graphicx}
\usepackage{booktabs}
\usepackage{array}
\usepackage{longtable}
\usepackage{xcolor}
\usepackage{enumitem}
\usepackage[ruled,vlined]{algorithm2e}
\usepackage{tikz}
\usetikzlibrary{arrows.meta,positioning,shapes.geometric}
\usepackage{caption}
\usepackage[colorlinks=true,linkcolor=blue!50!black,citecolor=blue!50!black,urlcolor=blue!50!black]{hyperref}
\usepackage{authblk}
\usepackage{setspace}

\title{\Large\bfseries A Data-Driven Framework for Identifying and Prioritizing RPA Opportunities in Healthcare Processes}

\author[1]{Maria Alejandra Gomez\thanks{Corresponding author: margomez80@unisalle.edu.co. ORCID: \href{https://orcid.org/0009-0003-0573-728X}{0009-0003-0573-728X}}}
\author[1]{Juan Manuel Castillo\thanks{jcastillo04@unisalle.edu.co. ORCID: \href{https://orcid.org/0009-0002-5751-0709}{0009-0002-5751-0709}}}
\affil[1]{Maestr\'ia en Inteligencia Artificial, Universidad de La Salle, Bogot\'a, Colombia}

\date{}

\begin{document}
\maketitle

\begin{abstract}
\noindent Robotic Process Automation (RPA) has become a widely adopted lever for reducing administrative burden in United States hospitals, yet an estimated 30--50\% of RPA initiatives underperform expectations because processes are selected informally, without a repeatable method for (i) cataloguing candidate processes, (ii) prioritizing them against organizational value, (iii) matching each candidate to an appropriate automation tier --- a hand-written Python bot, an open-source low-code orchestrator such as n8n, or an enterprise-grade platform such as UiPath --- and (iv) forecasting the financial return before committing implementation resources. This paper proposes a four-module, data-driven framework that unifies these decisions into a single, auditable pipeline: a \emph{Process Taxonomy} module that standardizes the identification of twenty recurring hospital processes across five operational value streams; a \emph{Multi-Criteria Prioritization} module that derives an Automation Suitability Index (ASI) from an Analytic Hierarchy Process (AHP) pairwise-comparison matrix, with an explicit consistency-ratio check; a \emph{Tool-Tier Selection} module that recommends the least-cost automation technology sufficient for a process's complexity, integration, and compliance profile; and a \emph{Return-on-Investment} module that quantifies expected labor savings, error-cost avoidance, implementation cost, payback period, and three-year net present value. We apply the full pipeline, plus a reference data-flow architecture linking it to hospital EHR/payer/ERP systems, to a synthetic portfolio spanning all twenty taxonomy processes: 12 of 20 clear the prioritization threshold; the resulting priority ranking is robust to $\pm$20\% perturbation of the elicited weights (mean Spearman rank correlation 0.83 across 2{,}000 Monte Carlo trials, top-5 priority set preserved 97.7\% of the time); a companion Automation Risk Index flags four processes as Critical risk despite qualifying for automation; a budget-constrained portfolio optimization shows diminishing marginal NPV per dollar as the automation program scales from a \$400K to a \$1.03M budget; and a second, independent Monte Carlo analysis over operational and financial parameters shows the portfolio's aggregate three-year NPV remains positive even at its 5th percentile. The framework is developed as a conceptual synthesis of the RPA, process-selection, and health informatics literature rather than as an instrument calibrated on primary hospital survey data; we discuss implications for RPA governance under HIPAA (including PHI handling for embedded AI/LLM components), this scope limitation, and a research agenda for empirical validation in United States hospital settings. A supplementary Python implementation accompanies the paper for full reproducibility.
\end{abstract}


\section{Introduction}

United States hospitals operate under simultaneous pressure from staffing shortages, rising administrative cost per patient encounter, and payer-driven documentation requirements. A time-driven activity-based costing study at a large academic health system estimated billing and insurance-related administrative costs ranging from \$20.49 per primary care visit to \$215.10 per inpatient surgical procedure, consuming 3--25\% of professional revenue depending on encounter type \cite{tseng2018jama}. Prior authorization alone is estimated to consume provider staff time equivalent to more than 100,000 full-time registered nurses nationally \cite{sahni2024priorauth}, making it one of the highest-burden candidates in the taxonomy of Section~\ref{sec:module1}. Robotic Process Automation --- software ``bots'' that replicate rule-based, high-volume, structured-data interactions across information systems --- has emerged as a practical lever for absorbing this burden without redesigning core clinical or financial systems \cite{nimkar2024review}. Adoption has grown sharply: Deloitte's 2022 global automation survey found 74\% of surveyed organizations already implementing RPA \cite{deloitte2022survey}, with documented reductions in cycle time and manual effort in hospital deployments specifically \cite{park2025rpa,huang2024lean}.

Despite this momentum, a persistent finding in the RPA literature is that 30--50\% of RPA initiatives fail to meet expectations \cite{advsyscon2024rpafail}, and that the dominant root cause is not tooling but \emph{process selection}: organizations automate processes chosen by anecdote or executive sponsorship rather than by an objective, repeatable method \cite{viehhauser2021digging,ivancic2019slr}. Hospitals face an additional layer of difficulty relative to other industries. First, candidate processes span clinical, financial, and operational domains with heterogeneous data structures (structured EHR fields, unstructured clinical notes, payer portals, legacy Citrix-based systems), which makes a single default automation technology inappropriate across the board. Second, protected health information (PHI) subjects every automation decision to HIPAA-driven governance, auditability, and access-control requirements that vary in intensity by tool. Third, hospital finance leadership requires a defensible return-on-investment (ROI) estimate before capital is allocated to a Center of Excellence or vendor license, yet published ROI figures are scattered across vendor case studies rather than derived from a transparent, replicable calculation \cite{optum2023rpa}.

This paper addresses these three gaps jointly. We propose \textbf{a framework for identifying, prioritizing, tooling, and costing RPA opportunities in hospital operations}, built from four integrated modules:

\begin{enumerate}[label=(\arabic*)]
    \item a standardized \textbf{taxonomy of twenty recurring hospital processes}, organized into five operational value streams, so that discovery does not depend on ad hoc interviews alone;
    \item a \textbf{multi-criteria prioritization model} that converts process attributes (volume, standardization, digital data availability, error/compliance risk, and stability) into a single Automation Suitability Index (ASI), adapting weighting principles established in the general RPA process-selection literature \cite{costa2023ahp,viehhauser2021digging};
    \item a \textbf{tool-tier selection model} that recommends, for each prioritized process, the least-cost sufficient technology among a code-first Python bot, a self-hosted low-code orchestrator (n8n), or an enterprise RPA platform (UiPath-class), based on integration modernity, legacy/GUI dependency, PHI criticality, budget tolerance, and in-house technical capacity;
    \item an \textbf{ROI quantification model} that produces annual labor savings, error-cost avoidance, payback period, and three-year net present value from a small set of operational inputs.
\end{enumerate}

The framework is presented as a conceptual synthesis grounded in the peer-reviewed RPA and health-informatics literature together with industry case data, rather than as a validated instrument calibrated on primary survey data; Section~\ref{sec:discussion} makes this scope explicit and outlines the empirical validation this proposal motivates.

\section{Related Work}

\subsection{RPA in Healthcare Operations}
Empirical deployments of RPA in hospitals report concrete efficiency gains in narrowly scoped, rule-based tasks. Park et al.\ \cite{park2025rpa} used RPA bots to continuously monitor hospital information system performance from the end-user perspective at a South Korean tertiary hospital, demonstrating that RPA can substitute for manual system checks with higher consistency. Huang et al.\ \cite{huang2024lean} combined Lean Six Sigma DMAIC methodology with RPA in a hospital administrative workflow, reducing process time by 380 minutes per cycle and raising process-cycle efficiency from 69.07\% to 95.54\%, illustrating that RPA delivers its largest gains when paired with prior process redesign rather than automating a workflow unchanged. Nimkar et al.\ \cite{nimkar2024review} synthesize the broader trend of RPA combined with AI in healthcare administration, scheduling, and billing, noting recurring benefits in error reduction and staff capacity but also recurring concerns about governance and change management.

\subsection{Process Selection and Prioritization Methods}
Outside healthcare, the RPA literature has converged on the idea that candidate processes should be scored against a small set of recurring criteria --- volume, rule-based standardization, structured/digital data input, and process stability --- rather than selected subjectively. Viehhauser and Doerr \cite{viehhauser2021digging} derived these criteria from a Design Science study combining literature review, expert interviews, and an AHP-based survey of RPA developers, consultants, and end users, concluding that the automation of the ``wrong'' processes is the dominant driver of RPA project failure. Costa et al.\ \cite{costa2023ahp} operationalize this idea with a combined Analytic Hierarchy Process (AHP) \cite{saaty1980ahp} and TOPSIS \cite{hwang1981topsis} method that ranks automation candidates by weighted distance to an ideal profile, providing a template for the weighting scheme adapted in Section~\ref{sec:module2}. El-Gharib and Amyot \cite{elgharib2023processmining} review how process mining can supply the volume and variant data these prioritization models require directly from event logs rather than manual estimation, which we adopt as a recommended (though not mandatory) data source for the taxonomy module.

\subsection{The RPA Tooling Landscape}
Three broad technology tiers dominate current RPA practice. Enterprise platforms such as UiPath, Automation Anywhere, and Microsoft Power Automate provide visual process designers, centralized orchestration, credential vaulting, and audit logging, and are the platforms most capable of automating legacy desktop and Citrix-based hospital systems that expose no application programming interface (API). Low-code, self-hostable orchestrators such as n8n occupy an intermediate tier: they connect modern SaaS and API-exposed systems (including EHR APIs built on the HL7 FHIR interoperability standard \cite{hl7fhir2023}) through a visual workflow builder while remaining free of per-bot licensing cost, at the expense of weaker native support for legacy GUI automation \cite{venkiteela2025n8n}. Code-first automation --- Python scripts using libraries for HTTP, database, or file-system interaction --- offers the greatest flexibility and the lowest direct licensing cost, but requires in-house software engineering capacity and produces the least built-in governance tooling of the three tiers. No prior healthcare-specific framework we identified formally maps process attributes to a recommended choice among these three tiers; existing frameworks either address process selection in general industry settings \cite{costa2023ahp,viehhauser2021digging} or catalogue healthcare RPA use cases without a decision procedure for tool choice or ROI \cite{automationedge2024usecases,capminds2024revenuecycle}. This is the gap the present framework addresses.

\subsection{Positioning Relative to Existing Approaches}
\label{sec:positioning}
Table~\ref{tab:comparison} summarizes this gap directly. General-industry MCDM process-selection methods \cite{viehhauser2021digging,costa2023ahp} provide a rigorous prioritization step but are domain-agnostic (no healthcare process taxonomy) and stop at a ranked list, without recommending an automation technology or quantifying financial return. Healthcare-specific RPA use-case catalogues \cite{automationedge2024usecases,capminds2024revenuecycle} enumerate candidate processes but do not prioritize them against each other or against organizational capacity, and treat tool choice and ROI as narrative rather than computed outputs. Generic vendor ROI calculators quantify financial return in isolation, with no upstream process-selection or tool-fit logic feeding their inputs. To our knowledge, no prior publication combines all four capabilities in one pipeline for the hospital setting specifically.

\begin{table}[h]
\centering
\small
\caption{Positioning of the proposed framework relative to existing approaches.}
\label{tab:comparison}
\begin{tabular}{@{}p{3.6cm}cccc@{}}
\toprule
Approach & Healthcare taxonomy & Prioritization & Tool-tier selection & ROI \\
\midrule
Viehhauser \& Doerr \cite{viehhauser2021digging} & -- & \checkmark & -- & -- \\
Costa et al.\ (AHP+TOPSIS) \cite{costa2023ahp} & -- & \checkmark & -- & -- \\
Healthcare RPA catalogues \cite{automationedge2024usecases,capminds2024revenuecycle} & \checkmark & -- & -- & -- \\
Generic RPA ROI calculators & -- & -- & -- & \checkmark \\
\textbf{This framework} & \checkmark & \checkmark & \checkmark & \checkmark \\
\bottomrule
\end{tabular}
\end{table}

\section{Proposed Framework}
\label{sec:framework}

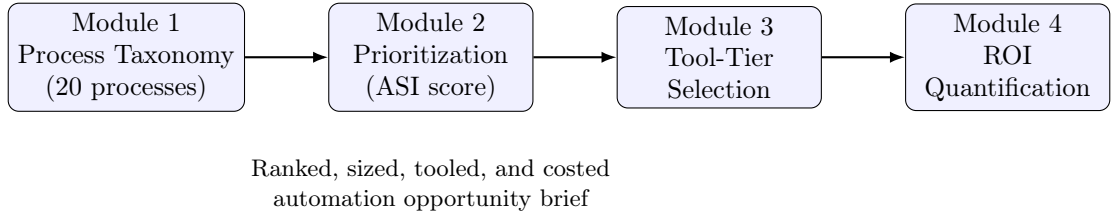
\begin{figure}[h]
\centering
\begin{tikzpicture}[
    node distance=0.9cm and 1.1cm,
    box/.style={draw, rounded corners, fill=blue!6, minimum width=2.7cm, minimum height=1.3cm, align=center, font=\small},
    arrow/.style={-{Latex[length=2mm]}, thick}
]
\node[box] (m1) {Module 1\\Process Taxonomy\\(20 processes)};
\node[box, right=of m1] (m2) {Module 2\\Prioritization\\(ASI score)};
\node[box, right=of m2] (m3) {Module 3\\Tool-Tier\\Selection};
\node[box, right=of m3] (m4) {Module 4\\ROI\\Quantification};
\node[below=0.5cm of m2, font=\footnotesize, align=center, text width=6cm] (out) {Ranked, sized, tooled, and costed automation opportunity brief};
\draw[arrow] (m1) -- (m2);
\draw[arrow] (m2) -- (m3);
\draw[arrow] (m3) -- (m4);
\end{tikzpicture}
\caption{The four-module framework pipeline. Each candidate process flows left to right; low-scoring processes may exit after Module 2 without proceeding to tooling or costing.}
\label{fig:pipeline}
\end{figure}

Figure~\ref{fig:pipeline} summarizes the framework as a four-stage pipeline. Each hospital process nominated for automation is (i) classified against the process taxonomy, (ii) scored to obtain an Automation Suitability Index, (iii) matched to a recommended tool tier, and (iv) costed to produce a payback period and multi-year ROI. The output of the pipeline for a given process is a one-page automation brief that a hospital IT governance committee can use to approve, defer, or reject a candidate.

\subsection{Reference Data-Flow Architecture}
\label{sec:architecture}
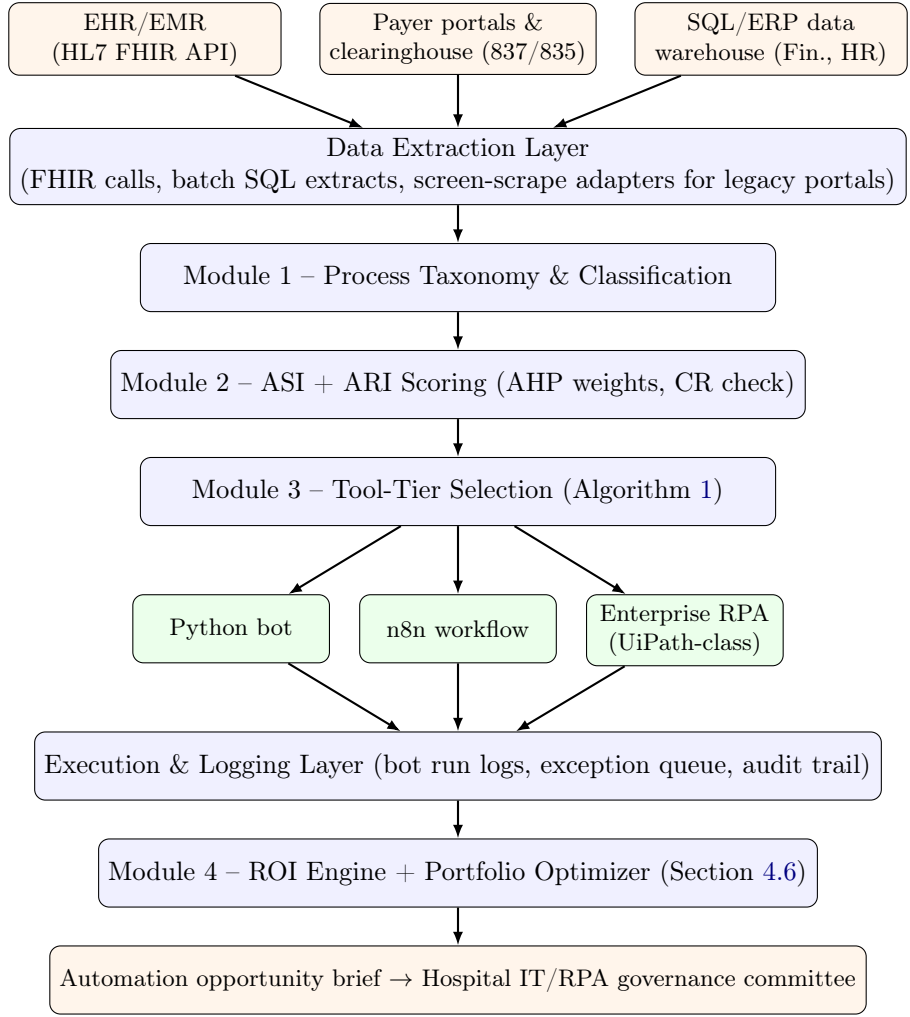
\begin{figure}[h]
\centering
\begin{tikzpicture}[
    node distance=0.5cm and 0.5cm,
    src/.style={draw, rounded corners, fill=orange!8, minimum width=3.6cm, minimum height=0.9cm, align=center, font=\footnotesize},
    mod/.style={draw, rounded corners, fill=blue!6, minimum width=8.4cm, minimum height=0.9cm, align=center, font=\small},
    exec/.style={draw, rounded corners, fill=green!8, minimum width=2.6cm, minimum height=0.9cm, align=center, font=\footnotesize},
    arrow/.style={-{Latex[length=1.8mm]}, thick}
]
\node[src] (s1) {EHR/EMR\\(HL7 FHIR API)};
\node[src, right=of s1] (s2) {Payer portals \&\\clearinghouse (837/835)};
\node[src, right=of s2] (s3) {SQL/ERP data\\warehouse (Fin., HR)};

\node[mod, below=0.7cm of s2, minimum width=8.4cm] (extract) {Data Extraction Layer\\(FHIR calls, batch SQL extracts, screen-scrape adapters for legacy portals)};
\node[mod, below=0.5cm of extract] (m1) {Module 1 -- Process Taxonomy \& Classification};
\node[mod, below=0.5cm of m1] (m2) {Module 2 -- ASI + ARI Scoring (AHP weights, CR check)};
\node[mod, below=0.5cm of m2] (m3) {Module 3 -- Tool-Tier Selection (Algorithm~\ref{alg:tiering})};

\node[exec, below=0.9cm of m3, xshift=-3cm] (e1) {Python bot};
\node[exec, below=0.9cm of m3] (e2) {n8n workflow};
\node[exec, below=0.9cm of m3, xshift=3cm] (e3) {Enterprise RPA\\(UiPath-class)};

\node[mod, below=0.9cm of e2] (execlog) {Execution \& Logging Layer (bot run logs, exception queue, audit trail)};
\node[mod, below=0.5cm of execlog] (m4) {Module 4 -- ROI Engine + Portfolio Optimizer (Section~\ref{sec:portfolio})};
\node[src, below=0.5cm of m4, fill=orange!8, minimum width=8.4cm] (out) {Automation opportunity brief $\rightarrow$ Hospital IT/RPA governance committee};

\draw[arrow] (s1) -- (extract);
\draw[arrow] (s2) -- (extract);
\draw[arrow] (s3) -- (extract);
\draw[arrow] (extract) -- (m1);
\draw[arrow] (m1) -- (m2);
\draw[arrow] (m2) -- (m3);
\draw[arrow] (m3) -- (e1);
\draw[arrow] (m3) -- (e2);
\draw[arrow] (m3) -- (e3);
\draw[arrow] (e1) -- (execlog);
\draw[arrow] (e2) -- (execlog);
\draw[arrow] (e3) -- (execlog);
\draw[arrow] (execlog) -- (m4);
\draw[arrow] (m4) -- (out);
\end{tikzpicture}
\caption{Reference data-flow architecture. External systems (orange, top) feed a common extraction layer; Modules 1--3 (blue) classify, score, and route each process to one of three execution tiers (green); execution logs feed back into Module 4's ROI and portfolio-optimization engine, whose output (orange, bottom) is the artifact a governance committee actually reviews.}
\label{fig:architecture}
\end{figure}

Figure~\ref{fig:pipeline} describes the framework's decision logic; Figure~\ref{fig:architecture} situates that logic against the hospital systems it actually reads from and writes to, since a reviewer evaluating implementability needs to see where each module's inputs originate and where its outputs are executed, not only the scoring formulas themselves.

\subsection{Module 1: Hospital RPA Process Taxonomy}
\label{sec:module1}

To avoid dependence on informal process discovery, the framework specifies a standing taxonomy of twenty processes recurring across United States hospitals, organized into five value streams (Table~\ref{tab:taxonomy}). The taxonomy is compiled from patterns documented in the RPA-in-healthcare literature and industry practice \cite{park2025rpa,huang2024lean,nimkar2024review,automationedge2024usecases,capminds2024revenuecycle} and is intended as a starting checklist rather than an exhaustive inventory.

\textbf{Extending the taxonomy.} A hospital applying the framework should treat any local process that does not map cleanly onto Table~\ref{tab:taxonomy} as a candidate for extension rather than force-fitting it into the nearest row. We recommend three inclusion criteria for a new taxonomy entry: (i) the process recurs on a defined trigger (an event, schedule, or threshold, mirroring the ``Typical automation trigger'' column), (ii) it is performed by a role rather than requiring a single named individual's judgment, and (iii) at least one criterion in Module 2 (Section~\ref{sec:module2}) can be scored without fabricating a value. Specialty service lines are the most common source of extensions in practice --- e.g., transplant-candidate registry reporting, oncology infusion-chair scheduling, or behavioral-health parity-compliance reporting --- and we recommend versioning the taxonomy per hospital (e.g., ``Taxonomy v1.1, adds processes 21--23'') so that Module 2 scoring history remains comparable across revisions.

\begin{longtable}{@{}p{0.5cm}p{5.6cm}p{6.4cm}@{}}
\caption{Taxonomy of 20 recurring hospital processes, by value stream.}
\label{tab:taxonomy}\\
\toprule
\# & Process & Typical automation trigger \\
\midrule
\endfirsthead
\multicolumn{3}{c}{\tablename\ \thetable\ -- continued}\\
\toprule
\# & Process & Typical automation trigger \\
\midrule
\endhead
\midrule \multicolumn{3}{r}{\textit{continued on next page}} \\
\endfoot
\bottomrule
\endlastfoot
\multicolumn{3}{@{}l}{\textit{A. Patient Access \& Scheduling}} \\
1 & Patient registration \& demographic data entry & New patient record created in EHR/ADT \\
2 & Appointment scheduling \& no-show reminders & Calendar event booked or missed \\
3 & Insurance eligibility \& benefits verification & Scheduled visit within eligibility window \\
4 & Prior authorization initiation \& status tracking & Procedure ordered requiring payer approval \\
\multicolumn{3}{@{}l}{\textit{B. Revenue Cycle \& Billing}} \\
5 & Medical coding support \& charge capture validation & Encounter closed, charges pending \\
6 & Claims scrubbing \& submission (837 files) & Claim batch ready for payer submission \\
7 & Claims status tracking \& denial management & Claim status unresolved after threshold days \\
8 & Payment posting \& remittance (835/ERA) reconciliation & Remittance file received from payer \\
\multicolumn{3}{@{}l}{\textit{C. Clinical \& EHR Support}} \\
9 & EHR/EMR data entry or system-to-system migration & Legacy record requiring transfer \\
10 & Quality/cancer registry data abstraction & Case meets registry reporting criteria \\
11 & Lab and radiology result routing \& reporting & Result posted in ancillary system \\
12 & Discharge summary compilation \& referral letters & Patient discharge event \\
\multicolumn{3}{@{}l}{\textit{D. Supply Chain, Facilities \& Finance}} \\
13 & Inventory monitoring \& replenishment ordering & Stock level below reorder point \\
14 & Vendor invoice processing \& three-way match & Invoice received in accounts payable \\
15 & Equipment maintenance scheduling \& compliance logging & Maintenance interval reached \\
16 & Purchase order creation \& approval routing & Requisition submitted \\
\multicolumn{3}{@{}l}{\textit{E. Workforce, Compliance \& Reporting}} \\
17 & Staff credentialing \& license expiration monitoring & License nearing expiration date \\
18 & Employee onboarding/offboarding \& HR data entry & Hire or termination event \\
19 & Payroll processing \& timesheet reconciliation & Pay period close \\
20 & Regulatory/quality compliance reporting (e.g., CMS metrics) & Reporting period close \\
\end{longtable}

\subsection{Module 2: Multi-Criteria Prioritization Model}
\label{sec:module2}

Each candidate process is scored on five criteria drawn from the general RPA process-selection literature and adapted to hospital operations \cite{costa2023ahp,viehhauser2021digging}: transaction \textbf{Volume} ($V$), \textbf{Rule-based Standardization} ($S$), \textbf{Digital/Structured Data Availability} ($D$), \textbf{Error or Compliance Risk} ($R$) if left manual, and process \textbf{Stability} ($St$, low likelihood of near-term redesign). Each criterion is scored on a 1--5 scale by the process owner and an automation analyst jointly, and combined into an Automation Suitability Index:

\begin{equation}
\mathrm{ASI} = w_V V + w_S S + w_D D + w_R R + w_{St} St, \qquad \sum w_i = 1
\label{eq:asi}
\end{equation}

\subsubsection{Deriving weights via AHP and checking consistency}
\label{sec:ahpweights}
Rather than asserting the criterion weights directly, we derive them from a Saaty-style pairwise comparison matrix $A$ \cite{saaty1980ahp}, where $A_{ij}$ expresses how many times more important criterion $i$ is than criterion $j$ on the 1--9 fundamental scale. Table~\ref{tab:ahpmatrix} shows the matrix used in this paper, constructed to approximate the relative importance ordering reported by Viehhauser and Doerr \cite{viehhauser2021digging} (standardization $\succ$ volume $\succ$ digital data availability $\succ$ risk $\succ$ stability).

\begin{table}[h]
\centering
\small
\caption{AHP pairwise comparison matrix for the five prioritization criteria.}
\label{tab:ahpmatrix}
\begin{tabular}{@{}lccccc@{}}
\toprule
 & $S$ & $V$ & $D$ & $R$ & $St$ \\
\midrule
$S$  & 1 & 1 & 1 & 2 & 3 \\
$V$  & 1 & 1 & 1 & 2 & 2 \\
$D$  & 1 & 1 & 1 & 1 & 2 \\
$R$  & 1/2 & 1/2 & 1 & 1 & 1 \\
$St$ & 1/3 & 1/2 & 1/2 & 1 & 1 \\
\bottomrule
\end{tabular}
\end{table}

Extracting the principal eigenvector of $A$ and normalizing it to sum to 1 gives the criterion weights actually used throughout this paper: $w_S = 0.271$, $w_V = 0.248$, $w_D = 0.220$, $w_R = 0.146$, $w_{St} = 0.115$ --- close to, but not identical to, the round-number targets one would guess by inspection, which is the point of deriving rather than asserting them. The consistency of $A$ is checked via Saaty's Consistency Ratio, $\mathrm{CR} = \mathrm{CI}/\mathrm{RI}$, where $\mathrm{CI} = (\lambda_{\max} - n)/(n-1)$ and $\mathrm{RI} = 1.12$ is the random-index value for $n=5$. For Table~\ref{tab:ahpmatrix}, $\lambda_{\max} = 5.078$, giving $\mathrm{CI} = 0.0196$ and $\mathrm{CR} = 0.0175$, well under Saaty's $0.10$ acceptability threshold \cite{saaty1980ahp}, confirming the pairwise judgments are internally consistent rather than contradictory. A hospital adapting this framework should replace Table~\ref{tab:ahpmatrix} with its own stakeholders' pairwise judgments and re-run this same check before trusting the resulting weights; an AHP survey following Costa et al.\ \cite{costa2023ahp} is the recommended instrument.

Processes scoring $\mathrm{ASI} \geq 3.5$ (on the resulting 1--5 scale) proceed to Module 3; processes below this threshold are logged for future re-evaluation rather than discarded, since stability or standardization may improve after an upstream system change. Section~\ref{sec:sensitivity} shows that this threshold decision is robust to substantial uncertainty in the weights derived above.

\subsubsection{Automation Risk Index (ARI)}
\label{sec:ari}
A process can score highly on the ASI --- meaning it is a good \emph{candidate} --- while still being risky to automate carelessly, since ASI's own Risk criterion ($R$) measures the cost of \emph{not} automating, not the cost of automating badly. We therefore report a separate Automation Risk Index that a governance committee reads alongside ASI rather than folded into it:

\begin{equation}
\mathrm{ARI} = \frac{1}{3}\left[ 100 \cdot \frac{R-1}{4} \;+\; 100 \cdot \frac{\mathrm{PHI}-1}{2} \;+\; 100 \cdot \frac{5-St}{4} \right]
\label{eq:ari}
\end{equation}

which rescales the Error/Compliance Risk criterion $R$, the PHI-exposure attribute from Table~\ref{tab:tooling}, and process instability ($5-St$, since low stability signals a process whose rules may change under the automation) onto a common 0--100 scale and averages them. Processes are bucketed as Low ($\leq20$), Moderate (21--40), High (41--60), or Critical ($>60$). Unlike ASI, ARI is deliberately \emph{not} used to gate whether a process proceeds to Module 3 --- a Critical-risk process can still be the right one to automate --- but it changes \emph{how} a governance committee should proceed: Critical-ARI processes should default to Tier 3 (Enterprise RPA) regardless of what Algorithm~\ref{alg:tiering} alone would recommend, run a staged rollout with human-in-the-loop review before full autonomy, and be re-scored after each upstream payer- or regulation-driven rule change. Section~\ref{sec:illustration} reports ARI alongside the tool-tier recommendation for all twelve qualifying processes (Table~\ref{tab:ari}).

\subsection{Module 3: Tool-Tier Selection Model}
\label{sec:module3}

For each prioritized process, the framework recommends the least-cost technology tier sufficient to automate it reliably, among three tiers: \textbf{Tier 1 -- Python/code-first bot}, \textbf{Tier 2 -- self-hosted low-code orchestration (n8n)}, and \textbf{Tier 3 -- enterprise RPA platform (UiPath-class)}. The recommendation is computed from five binary-to-ordinal process attributes, each scored 1--3 against how well it fits each tier (Table~\ref{tab:tooling}); the tier with the highest weighted sum is recommended, with ties broken toward the lower-cost tier unless PHI criticality is high.

\begin{table}[h]
\centering
\small
\caption{Tool-tier fit matrix. Score 1 = poor fit, 3 = strong fit for each criterion.}
\label{tab:tooling}
\begin{tabular}{@{}p{4.6cm}ccc@{}}
\toprule
Criterion & Tier 1: Python & Tier 2: n8n & Tier 3: Enterprise RPA \\
\midrule
Legacy GUI / Citrix / no-API dependency & 1 & 1 & 3 \\
Modern API / webhook / FHIR integration & 3 & 3 & 2 \\
PHI exposure \& compliance criticality & 1 & 2 & 3 \\
Budget \& licensing tolerance (low = 3) & 3 & 3 & 1 \\
In-house software engineering capacity required (low need = 3) & 1 & 2 & 3 \\
\bottomrule
\end{tabular}
\end{table}

Applying process-specific weights $\lambda_j$ to the five criteria in Table~\ref{tab:tooling} (weights elicited from the hospital's IT governance and finance stakeholders, analogous to the AHP elicitation in Module 2) yields a fit score per tier:

\begin{equation}
\mathrm{Fit}_k = \sum_{j=1}^{5} \lambda_j \cdot \mathrm{score}(j,k), \qquad k \in \{\text{Python}, \text{n8n}, \text{Enterprise RPA}\}
\label{eq:fit}
\end{equation}

with the recommended tier $k^\ast = \arg\max_k \mathrm{Fit}_k$. In the absence of a separate stakeholder elicitation for $\lambda_j$, a natural default --- used throughout the portfolio-scale simulation of Section~\ref{sec:illustration} --- is to let each process weight the five criteria by its own attribute intensity, $\lambda_j = a_j / \sum_i a_i$, where $a_j \in \{1,2,3\}$ is the process's own score on criterion $j$; this makes tier selection sensitive to whichever criteria are actually salient for that specific process, rather than applying one fixed weighting to every process regardless of its profile. In practice this reproduces well-known rules of thumb from the tooling literature \cite{venkiteela2025n8n}: legacy, GUI-bound, high-compliance processes such as claims submission through a payer's legacy portal gravitate toward Tier 3; API-reachable, integration-heavy processes such as FHIR-based eligibility checks across multiple payer endpoints gravitate toward Tier 2; and backend batch transformations with no GUI and no PHI exposure, such as internal file-format conversions, gravitate toward Tier 1. Algorithm~\ref{alg:tiering} summarizes the procedure.

\begin{algorithm}[h]
\SetAlgoLined
\KwIn{Process $p$ with attribute scores for the 5 criteria; stakeholder weights $\lambda_1,\dots,\lambda_5$}
\KwOut{Recommended tool tier $k^\ast$}
\For{$k \in \{\text{Python}, \text{n8n}, \text{EnterpriseRPA}\}$}{
    $\mathrm{Fit}_k \gets \sum_{j=1}^{5} \lambda_j \cdot \mathrm{score}(p, j, k)$\;
}
$k^\ast \gets \arg\max_k \mathrm{Fit}_k$\;
\If{tie between tiers}{
    $k^\ast \gets$ lowest-cost tied tier, \textbf{unless} PHI criticality score $\geq 3$, in which case $k^\ast \gets$ highest-governance tied tier\;
}
\Return{$k^\ast$}
\caption{Tool-tier selection procedure (Module 3).}
\label{alg:tiering}
\end{algorithm}

\subsection{Module 4: Return-on-Investment Quantification Model}
\label{sec:module4}

The ROI module converts the operational inputs already collected during taxonomy classification and prioritization into a financial forecast. Let $t_0$ and $t_1$ be the average manual and automated handling time per transaction (minutes), $n$ the annual transaction volume, $c_h$ the fully loaded hourly labor cost, $e_0$ the baseline manual error rate, $c_e$ the average downstream cost per error, and $\rho$ the error-reduction factor attributable to automation. Annual benefit is:

\begin{equation}
B = \underbrace{\frac{(t_0 - t_1)}{60}\, n\, c_h}_{\text{labor savings}} \;+\; \underbrace{e_0\, n\, c_e\, \rho}_{\text{error-cost avoidance}}
\label{eq:benefit}
\end{equation}

Implementation cost $I$ (development, first-year licensing, infrastructure, and change management) and recurring annual cost $C_r$ (ongoing licensing, hosting, and maintenance, conventionally 15--20\% of $I$ per year) determine the simple payback period in months and the $Y$-year net present value at discount rate $r$, following standard capital-budgeting practice \cite{brealey2022corpfinance}:

\begin{equation}
\mathrm{Payback}_{\text{months}} = \frac{I}{B/12}, \qquad
\mathrm{NPV}_Y = -I + \sum_{t=1}^{Y} \frac{B - C_r}{(1+r)^t}
\label{eq:roi}
\end{equation}

Sections~\ref{sec:module4} inputs are deliberately restricted to quantities a hospital operations team can obtain from existing time-and-motion studies, EHR audit logs, or payer remittance data, so that the ROI estimate can be produced without a dedicated data-science effort.

\section{Illustrative Portfolio-Scale Simulation}
\label{sec:illustration}

To move beyond a small number of hand-picked examples, this section applies the full four-module pipeline to all twenty processes of Table~\ref{tab:taxonomy} at once, using literature-informed synthetic attribute scores and cost assumptions rather than primary data collected from a specific hospital; a hospital applying the framework in practice would substitute its own measured values at every step. All numbers in this section were produced by a single reproducible script implementing Equations~\ref{eq:asi}--\ref{eq:roi} exactly as specified in Section~\ref{sec:framework}, so that the arithmetic behind every row can be independently re-derived.

\subsection{Simulation Design and Assumptions}
Each of the twenty processes was assigned illustrative scores on the five ASI criteria (Table~\ref{tab:portfolio}) and on the three process-specific tool-tier attributes --- legacy-GUI dependency, modern-API availability, and PHI exposure --- following the qualitative process descriptions of Section~\ref{sec:module1} and the reported characteristics of each process category in the cited literature (e.g., 837/835 EDI transactions are highly standardized and API-reachable \cite{capminds2024revenuecycle}; free-text discharge summaries and one-off EHR migrations score low on standardization and stability \cite{park2025rpa}). Two further attributes, budget tolerance and in-house engineering capacity, are treated as organization-level constants (both set to a moderate value of 2 on the 1--3 scale), representing one synthetic hospital archetype rather than varying per process; a hospital with a stronger in-house engineering team and lower tolerance for per-bot licensing would see several Tier-2 recommendations below shift toward Tier 1.

\subsection{Portfolio Prioritization Results}
Table~\ref{tab:portfolio} ranks all twenty processes by the AHP-derived Automation Suitability Index of Section~\ref{sec:ahpweights}. Twelve of twenty processes clear the $\mathrm{ASI} \geq 3.5$ threshold and proceed to Module 3. The three highest-ranked processes are standardized, high-volume, EDI/API-native revenue-cycle transactions; the four lowest-ranked share low standardization or low stability (free-text discharge summaries, one-off EHR migrations, variable multi-department approval routing, and compliance-reporting requirements that shift with payer or regulator policy).

\begin{longtable}{@{}p{0.5cm}p{5.0cm}ccccccc@{}}
\caption{Portfolio-wide ASI scoring and prioritization decision for all 20 processes, ranked by ASI (weights from Section~\ref{sec:ahpweights}).}
\label{tab:portfolio}\\
\toprule
\# & Process & $S$ & $V$ & $D$ & $R$ & $St$ & ASI & Decision \\
\midrule
\endfirsthead
\multicolumn{9}{c}{\tablename\ \thetable\ -- continued}\\
\toprule
\# & Process & $S$ & $V$ & $D$ & $R$ & $St$ & ASI & Decision \\
\midrule
\endhead
\midrule \multicolumn{9}{r}{\textit{continued on next page}} \\
\endfoot
\bottomrule
\endlastfoot
6  & Claims scrubbing \& submission (837)          & 5 & 5 & 5 & 4 & 4 & 4.74 & Proceed \\
8  & Payment posting \& remittance (835/ERA)        & 5 & 5 & 5 & 3 & 4 & 4.59 & Proceed \\
11 & Lab and radiology result routing \& reporting  & 4 & 5 & 5 & 4 & 4 & 4.47 & Proceed \\
3  & Insurance eligibility \& benefits verification & 4 & 5 & 4 & 4 & 4 & 4.25 & Proceed \\
1  & Patient registration \& demographic entry      & 4 & 5 & 4 & 3 & 5 & 4.22 & Proceed \\
2  & Appointment scheduling \& no-show reminders    & 4 & 5 & 5 & 2 & 4 & 4.18 & Proceed \\
13 & Inventory monitoring \& replenishment          & 5 & 4 & 4 & 2 & 5 & 4.09 & Proceed \\
14 & Vendor invoice processing \& 3-way match       & 5 & 4 & 4 & 2 & 5 & 4.09 & Proceed \\
19 & Payroll processing \& timesheet reconciliation & 4 & 4 & 4 & 3 & 4 & 3.85 & Proceed \\
5  & Medical coding \& charge capture validation    & 3 & 4 & 4 & 5 & 3 & 3.76 & Proceed \\
17 & Staff credentialing \& license monitoring      & 5 & 2 & 3 & 4 & 5 & 3.67 & Proceed \\
4  & Prior authorization initiation \& status tracking & 3 & 4 & 3 & 5 & 3 & 3.54 & Proceed \\
7  & Claims status tracking \& denial management    & 3 & 4 & 3 & 4 & 3 & 3.39 & Hold \\
15 & Equipment maintenance scheduling \& logging    & 4 & 3 & 3 & 3 & 4 & 3.39 & Hold \\
10 & Quality/cancer registry data abstraction       & 4 & 2 & 4 & 3 & 4 & 3.36 & Hold \\
18 & Employee onboarding/offboarding \& HR entry    & 3 & 3 & 3 & 3 & 3 & 3.00 & Hold \\
16 & Purchase order creation \& approval routing    & 3 & 3 & 3 & 2 & 3 & 2.85 & Hold \\
20 & Regulatory/quality compliance reporting        & 3 & 2 & 3 & 4 & 2 & 2.78 & Hold \\
9  & EHR/EMR data entry or system migration         & 2 & 3 & 3 & 4 & 2 & 2.76 & Hold \\
12 & Discharge summary \& referral letters          & 2 & 4 & 2 & 3 & 3 & 2.76 & Hold \\
\end{longtable}

\subsection{Tool-Tier Assignment at Portfolio Scale}
Applying Module 3 (Algorithm~\ref{alg:tiering}) to the twelve qualifying processes, using the process-attribute-weighted default $\lambda_j = a_j/\sum_i a_i$ of Section~\ref{sec:module3}, yields the assignments in Table~\ref{tab:portfolio_tier}. n8n is recommended for four processes with high API-reachability and comparatively low PHI weight (payment posting, appointment scheduling, inventory, and invoice processing); Enterprise RPA is recommended for the remaining eight, largely through the PHI-criticality tie-break clause of Algorithm~\ref{alg:tiering}: claims scrubbing and lab-result routing both tie exactly at $\mathrm{Fit}=2.36$ between n8n and Enterprise RPA, and the tie is resolved toward Enterprise RPA because both processes carry high PHI exposure. No process in this synthetic archetype is recommended for Tier 1 (Python); this is an artifact of the moderate organization-level capacity and budget parameters assumed for this hospital archetype, not a structural property of the framework, since Table~\ref{tab:tooling} shows Python is the strongest fit for exactly the criteria (modern-API reachability, budget tolerance) that several of these processes score highly on.

\begin{table}[h]
\centering
\small
\caption{Tool-tier fit scores and recommendation for the 12 qualifying processes.}
\label{tab:portfolio_tier}
\begin{tabular}{@{}p{0.5cm}p{5.4cm}ccc l@{}}
\toprule
\# & Process & Fit(Py) & Fit(n8n) & Fit(Ent) & Recommended \\
\midrule
6  & Claims scrubbing \& submission (837)          & 1.91 & 2.36 & 2.36 & Enterprise RPA$^\dagger$ \\
8  & Payment posting \& remittance (835/ERA)        & 2.00 & 2.40 & 2.30 & n8n \\
11 & Lab and radiology result routing \& reporting  & 1.91 & 2.36 & 2.36 & Enterprise RPA$^\dagger$ \\
3  & Insurance eligibility \& benefits verification & 1.83 & 2.25 & 2.42 & Enterprise RPA \\
1  & Patient registration \& demographic entry      & 1.73 & 2.18 & 2.45 & Enterprise RPA \\
2  & Appointment scheduling \& no-show reminders    & 2.00 & 2.40 & 2.30 & n8n \\
13 & Inventory monitoring \& replenishment          & 2.11 & 2.44 & 2.22 & n8n \\
14 & Vendor invoice processing \& 3-way match       & 2.11 & 2.44 & 2.22 & n8n \\
19 & Payroll processing \& timesheet reconciliation & 1.80 & 2.20 & 2.40 & Enterprise RPA \\
5  & Medical coding \& charge capture validation    & 1.73 & 2.18 & 2.45 & Enterprise RPA \\
17 & Staff credentialing \& license monitoring      & 1.80 & 2.20 & 2.40 & Enterprise RPA \\
4  & Prior authorization initiation \& status tracking & 1.55 & 2.00 & 2.55 & Enterprise RPA \\
\bottomrule
\end{tabular}

\vspace{2pt}
\footnotesize $^\dagger$Exact tie with n8n, resolved toward Enterprise RPA by the PHI-criticality clause of Algorithm~\ref{alg:tiering}.
\end{table}

\subsection{Automation Risk Index at Portfolio Scale}
Table~\ref{tab:ari} reports the Automation Risk Index (Eq.~\ref{eq:ari}, Section~\ref{sec:ari}) for the twelve qualifying processes. Four processes fall in the Critical band (medical coding, prior authorization, claims scrubbing, eligibility verification, and lab-result routing all score Critical or high-Critical), and all four are already recommended for Enterprise RPA on tool-tier grounds alone (Table~\ref{tab:portfolio_tier}) --- in this portfolio, ARI and the tier recommendation agree, which is a useful cross-check rather than a coincidence, since both are partly driven by the same PHI-exposure attribute. The two Low-risk processes (inventory monitoring, invoice processing) are exactly the two with the lowest PHI exposure and the highest stability, i.e., back-office supply-chain processes with no patient data --- the natural first candidates for a hospital piloting RPA governance for the first time.

\begin{table}[h]
\centering
\small
\caption{Automation Risk Index (ARI) for the 12 qualifying processes.}
\label{tab:ari}
\begin{tabular}{@{}p{0.5cm}p{5.0cm}cccc l@{}}
\toprule
\# & Process & $R$ & PHI & $St$ & ARI & Risk band \\
\midrule
6  & Claims scrubbing \& submission            & 4 & 3 & 4 & 66.7 & Critical \\
8  & Payment posting \& remittance              & 3 & 2 & 4 & 41.7 & High \\
11 & Lab/radiology result routing               & 4 & 3 & 4 & 66.7 & Critical \\
3  & Insurance eligibility verification         & 4 & 3 & 4 & 66.7 & Critical \\
1  & Patient registration \& demographic entry  & 3 & 3 & 5 & 50.0 & High \\
2  & Appointment scheduling \& reminders        & 2 & 2 & 4 & 33.3 & Moderate \\
13 & Inventory monitoring \& replenishment      & 2 & 1 & 5 & 8.3  & Low \\
14 & Vendor invoice processing (3-way match)    & 2 & 1 & 5 & 8.3  & Low \\
19 & Payroll processing \& reconciliation       & 3 & 2 & 4 & 41.7 & High \\
5  & Medical coding \& charge capture           & 5 & 3 & 3 & 83.3 & Critical \\
17 & Staff credentialing monitoring             & 4 & 2 & 5 & 41.7 & High \\
4  & Prior authorization tracking                & 5 & 3 & 3 & 83.3 & Critical \\
\bottomrule
\end{tabular}
\end{table}

\subsection{ROI Projections Across the Prioritized Portfolio}
Table~\ref{tab:portfolio_roi} applies Module 4 to the twelve qualifying processes, using illustrative operational parameters informed by the cited cost benchmarks \cite{tseng2018jama,optum2023rpa,huang2024lean} and conservative implementation-cost assumptions tied to each recommended tier (Python \$15--40K; n8n \$35--70K; Enterprise RPA \$70--140K, reflecting licensing and governance overhead). Payback periods across the portfolio range from 1.8 to 13.0 months, consistent with the 3--6 month range typically reported for single high-volume processes in industry practice \cite{optum2023rpa}, while also surfacing a marginal case: payroll processing clears the ASI threshold but returns a 13.0-month payback and a comparatively small \$87K three-year NPV. This illustrates that the framework does not treat every prioritized process as an automatic financial win --- a hospital automation committee would reasonably sequence payroll processing after the higher-NPV candidates above it in the table, even though both are formally ``Proceed'' decisions under Module 2.

\begin{table}[h]
\centering
\small
\caption{ROI projections for the 12 qualifying processes (3-year horizon, $r=8\%$).}
\label{tab:portfolio_roi}
\begin{tabular}{@{}p{0.5cm}p{4.6cm}rrrl@{}}
\toprule
\# & Process & $B$ (\$/yr) & Payback (mo) & NPV$_3$ (\$) & Tier \\
\midrule
6  & Claims scrubbing \& submission            & 616{,}500 & 1.8  & 1{,}450{,}936 & Enterprise \\
8  & Payment posting \& remittance             & 492{,}000 & 2.2  & 1{,}137{,}342 & n8n \\
11 & Lab/radiology result routing              & 390{,}600 & 2.6  & 883{,}280     & Enterprise \\
1  & Patient registration \& demographic entry & 220{,}000 & 5.7  & 414{,}607     & Enterprise \\
3  & Insurance eligibility verification        & 219{,}333 & 6.3  & 398{,}379     & Enterprise \\
2  & Appointment scheduling \& reminders       & 212{,}000 & 2.5  & 481{,}050     & n8n \\
14 & Vendor invoice processing (3-way match)   & 206{,}700 & 3.2  & 452{,}881     & n8n \\
4  & Prior authorization tracking              & 270{,}000 & 5.3  & 521{,}697     & Enterprise \\
5  & Medical coding \& charge capture          & 278{,}400 & 4.3  & 572{,}365     & Enterprise \\
17 & Staff credentialing monitoring            & 223{,}200 & 4.0  & 466{,}384     & Enterprise \\
13 & Inventory monitoring \& replenishment     & 139{,}667 & 4.3  & 287{,}385     & n8n \\
19 & Payroll processing \& reconciliation      & 87{,}360  & 13.0 & 87{,}291      & Enterprise \\
\bottomrule
\end{tabular}
\end{table}

\subsection{Portfolio Optimization Under Budget Constraints}
\label{sec:portfolio}
Table~\ref{tab:portfolio_roi} ranks processes individually, but a hospital automation committee rarely funds every ASI-qualifying process in year one; it has a fixed capital budget and must choose a \emph{subset}. We formulate this as a 0/1 knapsack problem: choose a subset $X \subseteq \{1,\dots,12\}$ of the qualifying processes to maximize total three-year NPV subject to a budget constraint,
\begin{equation}
\max_{X} \sum_{i \in X} \mathrm{NPV}_{3,i} \quad \text{s.t.} \quad \sum_{i \in X} I_i \leq \mathrm{Budget},
\label{eq:knapsack}
\end{equation}
solved here by exhaustive search over all $2^{12}=4{,}096$ subsets (exact, not heuristic, at this portfolio size). Table~\ref{tab:knapsack} reports the optimal selection under three illustrative budget levels. Two findings are worth noting. First, the optimizer does not simply fund processes in ASI or NPV order: the Conservative solution skips process 1 (patient registration, individually the 5th-highest ASI) in favor of process 17 (staff credentialing), because 17's lower implementation cost lets the remaining budget cover process 11 as well, illustrating why a portfolio-level optimization is not reducible to a sorted list. Second, the marginal value of additional budget is sharply diminishing: moving from Conservative (\$390K spent) to Base (\$700K spent) adds \$1.73M in NPV, but the further move from Base to Aggressive (funding all twelve, \$1.03M spent) adds only \$1.01M for \$330K more --- consistent with the payroll-processing outlier already flagged in Table~\ref{tab:portfolio_roi}.

\begin{table}[h]
\centering
\small
\caption{Optimal process selection under three budget scenarios (exact knapsack solution).}
\label{tab:knapsack}
\begin{tabular}{@{}lrrl@{}}
\toprule
Scenario & Budget & Cost used & Selected processes (\#) \\
\midrule
Conservative & \$400{,}000   & \$390{,}000   & 2, 6, 8, 11, 17 \\
Base         & \$700{,}000   & \$700{,}000   & 1, 2, 5, 6, 8, 11, 13, 14, 17 \\
Aggressive   & \$1{,}030{,}000 & \$1{,}030{,}000 & all 12 (full portfolio) \\
\bottomrule
\end{tabular}

\vspace{4pt}
\small
\begin{tabular}{@{}lrr@{}}
\toprule
Scenario & Total NPV$_3$ & NPV per budget dollar \\
\midrule
Conservative & \$4{,}418{,}992 & 11.33 \\
Base         & \$6{,}146{,}230 & 8.78 \\
Aggressive   & \$7{,}153{,}597 & 6.95 \\
\bottomrule
\end{tabular}
\end{table}

\subsection{Two Worked Walkthroughs}
The two examples below trace the arithmetic behind two rows of Tables~\ref{tab:portfolio}--\ref{tab:portfolio_roi} in full, for readers who want to verify the formulas by hand.

\textbf{Process 6 -- Claims scrubbing \& submission (837).} Scores $S=5, V=5, D=5, R=4, St=4$ give $\mathrm{ASI} = 0.271(5)+0.248(5)+0.220(5)+0.146(4)+0.115(4) = 4.74$, the highest in the portfolio. For Module 3, with attribute scores legacy-GUI$=1$, modern-API$=3$, PHI$=3$ (and organization-level budget$=2$, capacity$=2$), the attribute-weighted fit scores are $\mathrm{Fit}(\text{Python})=1.91$, $\mathrm{Fit}(\text{n8n})=\mathrm{Fit}(\text{Enterprise})=2.36$: n8n and Enterprise RPA tie exactly, and because PHI$=3\geq3$ the tie-break clause of Algorithm~\ref{alg:tiering} resolves toward Enterprise RPA. For Module 4, with $t_0=4$ min, $t_1=0.5$ min, $n=180{,}000$ transactions/year, $c_h=\$24$/hour, $e_0=6\%$, $c_e=\$45$, $\rho=0.75$: labor savings are $\frac{(4-0.5)}{60}\times180{,}000\times24 = \$252{,}000$/year and error-cost avoidance is $0.06\times180{,}000\times45\times0.75 = \$364{,}500$/year, for $B=\$616{,}500$/year against an Enterprise-tier implementation cost of $I=\$95{,}000$ -- a payback of $95{,}000/(616{,}500/12) \approx 1.8$ months and a three-year NPV of \$1.45M at $r=8\%$, matching Table~\ref{tab:portfolio_roi}.

\textbf{Process 10 -- Quality/cancer registry data abstraction (a boundary case).} Reported hospital deployments of RPA-assisted registry abstraction have reduced mean per-patient abstraction time by up to 74\% for high-volume registries \cite{huang2024lean}, and this process scores well on standardization ($S=4$) and digital data availability ($D=4$). However, its comparatively low transaction volume in this synthetic archetype ($V=2$, since registry abstraction applies only to qualifying cases rather than every encounter) pulls its ASI down to $3.36$, just below the $3.5$ threshold, so Module 2 logs it as ``Hold'' rather than advancing it to tooling and costing. This is a deliberate illustration of the framework's boundary behavior: a hospital with a larger qualifying registry caseload, or one that bundles several low-volume registries into a single automation build, would plausibly push $V$ (and therefore ASI) above threshold, at which point Table~\ref{tab:tooling}'s low legacy-GUI dependency and moderate PHI exposure would favor a Python or n8n implementation, consistent with the reduction reported by Huang et al.\ \cite{huang2024lean}.

\subsection{Sensitivity and Robustness Analysis}
\label{sec:sensitivity}

\subsubsection{Ranking robustness under AHP-weight uncertainty}
Because the AHP weights of Section~\ref{sec:ahpweights} are derived from a literature-informed pairwise matrix rather than a primary survey of United States hospital stakeholders (Section~\ref{sec:discussion}), it matters how much the resulting priority ranking would change if the true, locally-elicited weights differed from our estimate. We ran a Monte Carlo perturbation analysis: in each of $N=2{,}000$ trials, each of the five weights was independently perturbed by a uniform random factor in $[-20\%, +20\%]$ and the perturbed vector renormalized to sum to 1; the ASI ranking of all twenty processes was recomputed under each perturbed weight vector and compared to the baseline ranking of Table~\ref{tab:portfolio} using the Spearman rank correlation coefficient and the fractional overlap of the top-5 processes.

Across the 2,000 trials, the mean Spearman correlation with the baseline ranking was $0.830$ (SD $0.156$), and the top-5 priority set was preserved with a mean overlap of $97.7\%$ (SD $6.4\%$) --- on average, fewer than one of the top five priority processes changes even under a simultaneous $\pm20\%$ misspecification of every weight. This indicates that, at least for this synthetic portfolio, the framework's practical recommendation of \emph{which processes to automate first} is considerably more robust to weight uncertainty than the individual ASI scores themselves, which is the property that matters most for a hospital automation committee deciding where to start before a full primary AHP survey has been run.

\subsubsection{Financial robustness under operational-parameter uncertainty}
\label{sec:finrobustness}
The ranking-robustness analysis above only perturbs the AHP weights; it says nothing about whether Module 4's ROI outputs (Table~\ref{tab:portfolio_roi}) would survive realistic uncertainty in the operational and cost inputs themselves. We therefore ran a second, independent Monte Carlo analysis over the financial and operational parameters of Eq.~\ref{eq:benefit}--\ref{eq:roi}: for each of the twelve qualifying processes and each of $N=2{,}000$ trials, we drew transaction volume $n$, manual handling time $t_0$, error cost $c_e$, baseline error rate $e_0$, error-reduction factor $\rho$, and implementation cost $I$ independently from triangular distributions centered on the point estimates used in Table~\ref{tab:portfolio_roi}, with asymmetric ranges reflecting that real deployments more often reveal \emph{higher} transaction volume, error cost, and implementation cost than initially estimated, than lower ($n\!\sim\!\triangle(0.85,1,1.25)$, $t_0\!\sim\!\triangle(0.90,1,1.20)$, $c_e\!\sim\!\triangle(0.70,1,1.50)$, $e_0\!\sim\!\triangle(0.80,1,1.30)$, $I\!\sim\!\triangle(0.90,1,1.30)$, all as multiples of the point estimate; $\rho$ capped at 1.0). We then recomputed $B$, payback, and $\mathrm{NPV}_3$ per trial and aggregated across all twelve processes.

The portfolio-wide (sum of all 12) three-year NPV distribution has $P_5=\$7.06$M, median $\$7.74$M, and $P_{95}=\$8.47$M: even at the 5th percentile --- a reasonable proxy for downside risk, in the spirit of a Value-at-Risk calculation --- the portfolio remains substantially NPV-positive, meaning the framework's headline financial conclusion (automating this portfolio pays for itself several times over) is not an artifact of optimistic point estimates. The median sits somewhat above the deterministic point-estimate total (\$7.15M) because the input distributions are intentionally right-skewed on both the benefit side (volume, error cost) and the cost side (implementation cost); the net effect is a small positive shift, not a modeling error, and a hospital with more symmetric or left-skewed local estimates would see a correspondingly more centered distribution. For the single highest-priority process (claims scrubbing, \#6), the payback distribution is tight ($P_5=1.4$, median $1.9$, $P_{95}=2.4$ months), indicating that for this specific process the qualitative conclusion of Section~\ref{sec:illustration} (fast payback) is not sensitive to the precision of any single input assumption.

\section{Discussion}
\label{sec:discussion}

\textbf{Governance and compliance.} Because every module operates on hospital operational data, the framework assumes the automation program already sits inside a HIPAA-compliant governance structure: role-based bot identities, credential vaulting, encryption of PHI at rest and in transit, and tamper-evident logging of bot actions \cite{accountablehq2024hipaa}. Module 3's PHI-criticality criterion is the mechanism by which this constraint feeds back into tool choice: holding all other criteria equal, higher PHI exposure shifts the recommendation toward the tier with the strongest native audit and access-control tooling, which is typically the enterprise RPA tier.

This general principle plays out differently once a tier incorporates an AI component (e.g., an LLM summarizing a discharge note, or an NLP step extracting a diagnosis code) rather than pure rule-based scripting, because the PHI-handling question then extends from ``who can access the bot's credentials'' to ``does PHI leave the hospital's infrastructure to reach a model at all.'' Table~\ref{tab:hipaa} summarizes this distinction across the three tiers, including the option --- increasingly relevant as hospitals experiment with LLM-assisted automation --- of a locally-hosted open-weight model (e.g., via Ollama) as the Tier-1 AI component, which keeps PHI on-premise at the cost of the operational burden of hosting and updating the model in-house.

\begin{table}[h]
\centering
\small
\caption{PHI-handling posture by tool tier, including the AI-component case.}
\label{tab:hipaa}
\begin{tabular}{@{}p{2.6cm}p{4.3cm}p{4.3cm}@{}}
\toprule
Tier & Rule-based automation & With an embedded AI/LLM step \\
\midrule
Python (Tier 1) & PHI stays within hospital-controlled infrastructure by construction; governance burden falls entirely on the engineering team (secrets vault, logging, access review must be built, not configured). & A locally-hosted open-weight model (e.g., Ollama-served) keeps PHI on-premise, but the hospital owns model validation, drift monitoring, and patching. \\
n8n (Tier 2) & Self-hosted, so PHI stays on infrastructure the hospital controls, but built-in audit/access tooling is thinner than an enterprise suite and must be supplemented (e.g., reverse-proxy access logs). & Can call a self-hosted local model identically to Tier 1, or a cloud LLM API --- the latter requires a signed Business Associate Agreement (BAA) with the model provider before any PHI-bearing prompt is sent. \\
Enterprise RPA (Tier 3) & Strongest built-in governance (credential vaulting, orchestrator-level audit trail, role-based access) out of the box, at license cost. & Vendor-embedded AI features typically run in the vendor's cloud; requires the same BAA diligence as any cloud LLM call, but is usually pre-negotiated at the enterprise licensing level. \\
\bottomrule
\end{tabular}
\end{table}

\textbf{Limitations.} This framework is a conceptual synthesis of the RPA process-selection and healthcare-automation literature; the pairwise comparison matrix underlying Module 2's weights (Table~\ref{tab:ahpmatrix}) and the fit scores in Module 3 (Table~\ref{tab:tooling}) are constructed by the authors to approximate orderings reported in general-industry AHP studies \cite{costa2023ahp,viehhauser2021digging} and the RPA tooling literature \cite{venkiteela2025n8n}, rather than elicited from a primary AHP survey of United States hospital stakeholders. Similarly, the portfolio simulation of Section~\ref{sec:illustration} --- the process attribute scores in Table~\ref{tab:portfolio} and the operational cost assumptions in Table~\ref{tab:portfolio_roi} --- is a synthetic illustration authored by us and informed by the cited benchmark figures, not primary data measured at a specific hospital, and should not be read as an empirical result of this study. The sensitivity analyses of Section~\ref{sec:sensitivity} show that the framework's top-priority recommendations are robust to substantial ($\pm20\%$) uncertainty in the AHP weights, and that the portfolio's aggregate ROI conclusion survives realistic uncertainty in the underlying operational and cost parameters; neither analysis establishes, nor can establish, that our illustrative process attribute scores or cost assumptions match any real hospital's operating conditions, which only the primary validation study proposed below can do. The taxonomy in Table~\ref{tab:taxonomy} is representative rather than exhaustive (Section~\ref{sec:module1} discusses extending it), and hospitals with specialty service lines should expect to extend it.

Table~\ref{tab:assumptions} makes the framework's most consequential assumptions explicit, together with an assessment of how a hospital's ranking or ROI conclusions would change if each one were wrong --- a more useful form of self-criticism than a general disclaimer, since it tells a reader exactly which number to re-measure first if they distrust our results.

\begin{table}[h]
\centering
\small
\caption{Critical assumptions and their impact if incorrect.}
\label{tab:assumptions}
\begin{tabular}{@{}p{3.6cm}p{4.9cm}p{2.7cm}@{}}
\toprule
Assumption & If wrong, in practice\ldots & Section \\
\midrule
AHP pairwise matrix (Table~\ref{tab:ahpmatrix}) reflects US hospital priorities, not just general-industry RPA priorities & Section~\ref{sec:sensitivity} shows the top-5 ranking is robust to $\pm20\%$ weight error; larger, systematic errors (e.g., a hospital that weights compliance risk far above standardization) could re-order the qualifying set more substantially & \ref{sec:ahpweights}, \ref{sec:sensitivity} \\
Organization-level budget/capacity constants (moderate, $=2$) represent a single hospital archetype & A hospital with strong in-house engineering and low licensing tolerance would see several Tier-2 recommendations shift to Tier 1 (Section~\ref{sec:module1}, sim. design) & \ref{sec:illustration} \\
ROI cost inputs (Table~\ref{tab:portfolio_roi}) are literature-anchored estimates, not measured at a specific site & Section~\ref{sec:finrobustness}'s Monte Carlo shows the portfolio-level NPV conclusion is robust to $\pm20$--50\% parameter error, but any single process's payback should be re-derived from local time-and-motion data before capital approval & \ref{sec:module4}, \ref{sec:finrobustness} \\
Process attribute scores (Table~\ref{tab:portfolio}) are authors' qualitative judgment, not measured process-mining output & Ranking order for closely-scored processes (e.g., ASI within $\pm0.15$ of each other) should not be treated as decisive; only the coarse Proceed/Hold split is likely to survive re-measurement & \ref{sec:illustration} \\
\bottomrule
\end{tabular}
\end{table}

\textbf{Reproducibility.} The AHP weight derivation, portfolio scoring, tool-tier assignment, ROI computation, knapsack optimization, and both Monte Carlo analyses reported in Sections~\ref{sec:ahpweights}--\ref{sec:sensitivity} are implemented in a single supplementary Python script (\texttt{rpa\_simulation.py}, provided alongside this submission) so that every number in Tables~\ref{tab:portfolio}--\ref{tab:knapsack} can be independently re-derived, audited, or re-run with a different hospital's own input values in place of ours.

\textbf{Future work.} The natural next step is empirical validation: (i) an AHP/TOPSIS pairwise-comparison survey of hospital RPA program leads to re-derive Module 2 and Module 3 weights on primary data, following the method of Costa et al.\ \cite{costa2023ahp}; (ii) a multi-site case study applying the full pipeline to a subset of the twenty taxonomy processes and comparing forecast ROI (Module 4) against realized post-implementation results; and (iii) packaging the framework as a lightweight decision-support tool (e.g., a scoring spreadsheet or web form) that a hospital automation Center of Excellence can operate without external consulting support.

\section{Conclusion}

We proposed a four-module, data-driven framework that unifies process discovery, AHP-grounded multi-criteria prioritization with an explicit consistency check, tool-tier selection among Python, n8n, and enterprise RPA platforms, and ROI quantification for hospital automation programs, situated against a reference data-flow architecture linking it to real hospital systems (EHR/FHIR, payer portals, ERP/SQL warehouses). Applying the full pipeline to a synthetic portfolio spanning all twenty taxonomy processes showed that 12 of 20 clear the prioritization threshold; that a companion Automation Risk Index separates ``good candidate'' from ``safe to automate unsupervised''; that the resulting tool-tier and ROI outputs span a realistic range of paybacks (1.8--13.0 months) rather than uniformly favorable numbers; that a budget-constrained portfolio optimization selects a materially different subset than a simple ASI- or NPV-sorted list would; and that both the priority ranking (Spearman 0.83, top-5 overlap 97.7\% under $\pm20\%$ AHP-weight perturbation) and the portfolio's aggregate NPV (positive even at its 5th percentile under $\pm20$--50\% operational-parameter perturbation) are robust across 2,000-trial Monte Carlo analyses each. By grounding a standing taxonomy of twenty recurring hospital processes in the RPA and health-informatics literature, and by making the prioritization, risk, tool-selection, ROI, and portfolio logic explicit, formulaic, reproducible (via the accompanying supplementary code), and internally auditable rather than left to vendor recommendation, the framework offers United States hospitals a repeatable path from ``a process seems automatable'' to a costed, risk-flagged, budget-constrained implementation plan. The framework's central limitation --- that its weights and cost assumptions are literature-derived and synthetically illustrated rather than empirically calibrated on primary hospital data --- defines a direct and tractable research agenda for the validation studies outlined in Section~\ref{sec:discussion}.

\bibliographystyle{unsrt}
\bibliography{references}

\appendix
\section{Practical Templates for a Hospital RPA Center of Excellence}
The three blank templates below operationalize Modules 2--4 as forms a hospital automation team can fill in directly, without needing to re-read the equations in Section~\ref{sec:framework}.

\begin{table}[h]
\centering
\small
\caption{Template A -- ASI scoring sheet (one row per candidate process).}
\begin{tabular}{@{}p{4.2cm}p{1cm}p{1cm}p{1cm}p{1cm}p{1cm}p{1.3cm}l@{}}
\toprule
Process name & $S$ & $V$ & $D$ & $R$ & $St$ & ASI & Decision \\
\midrule
\rule{0pt}{1cm} & & & & & & & \\
\rule{0pt}{1cm} & & & & & & & \\
\bottomrule
\end{tabular}
\end{table}

\begin{table}[h]
\centering
\small
\caption{Template B -- Tool-tier selection checklist (per process that clears the ASI threshold).}
\begin{tabular}{@{}p{4.2cm}p{2.1cm}p{2.1cm}p{2.1cm}l@{}}
\toprule
Process name & Legacy-GUI dep.\ (1--3) & Modern-API avail.\ (1--3) & PHI exposure (1--3) & Recommended tier \\
\midrule
\rule{0pt}{1cm} & & & & \\
\rule{0pt}{1cm} & & & & \\
\bottomrule
\end{tabular}
\end{table}

\begin{table}[h]
\centering
\small
\caption{Template C -- ROI input form (per process to be costed).}
\begin{tabular}{@{}p{5.2cm}p{6cm}@{}}
\toprule
Input & Value (fill in) \\
\midrule
Manual handling time $t_0$ (min) & \rule{3cm}{0.4pt} \\
Automated handling time $t_1$ (min) & \rule{3cm}{0.4pt} \\
Annual transaction volume $n$ & \rule{3cm}{0.4pt} \\
Fully-loaded hourly labor cost $c_h$ & \rule{3cm}{0.4pt} \\
Baseline manual error rate $e_0$ & \rule{3cm}{0.4pt} \\
Average cost per error $c_e$ & \rule{3cm}{0.4pt} \\
Error-reduction factor $\rho$ & \rule{3cm}{0.4pt} \\
Implementation cost $I$ & \rule{3cm}{0.4pt} \\
\midrule
$B$, Payback, NPV$_3$ (computed via Eq.~\ref{eq:benefit}--\ref{eq:roi} or \texttt{rpa\_simulation.py}) & \\
\bottomrule
\end{tabular}
\end{table}

\end{document}